\documentclass[conference]{IEEEtran}
\IEEEoverridecommandlockouts
\usepackage{cite}
\usepackage{amsmath,amssymb,amsfonts}
\usepackage{algorithmic}
\usepackage{graphicx}
\usepackage{textcomp}
\usepackage{xcolor}

\newtheorem{definition}{Definition}[section]

\newtheorem{lemma}{Lemma}[section]

\def\BibTeX{{\rm B\kern-.05em{\sc i\kern-.025em b}\kern-.08em
    T\kern-.1667em\lower.7ex\hbox{E}\kern-.125emX}}

\makeatletter
\newcommand{\linebreakand}{%
  \end{@IEEEauthorhalign}
  \hfill\mbox{}\par
  \mbox{}\hfill\begin{@IEEEauthorhalign}
}
\makeatother

\begin{document}

\title{Position Paper: Rethinking Privacy in RL for Sequential Decision-making in the Age of LLMs
}

\author{
\IEEEauthorblockN{Flint Xiaofeng Fan}
\IEEEauthorblockA{\textit{Centre for Frontier AI Research (CFAR)} \\
\textit{Institute of High Performance Computing (IHPC)} \\
\textit{Agency for Science, Technology and Research }\\
Singapore \\
fxf@u.nus.edu}
\and
\IEEEauthorblockN{Cheston Tan}
\IEEEauthorblockA{\textit{Centre for Frontier AI Research (CFAR)} \\
\textit{Institute of High Performance Computing (IHPC)} \\
\textit{Agency for Science, Technology and Research }\\
Singapore \\
cheston-tan@i2r.a-star.edu.sg}

\linebreakand 

\IEEEauthorblockN{Roger Wattenhofer}
\IEEEauthorblockA{\textit{D-ITET} \\
\textit{ETH Zurich}\\
Zurich, Switzerland \\
wattenhofer@ethz.ch}
\and


\IEEEauthorblockN{Yew-Soon Ong}
\IEEEauthorblockA{
\textit{CCDS, Nanyang Technological University}\\
\textit{CFAR, IHPC, Agency for Science, Technology and Research }\\
Singapore \\
asysong@ntu.edu.sg}
}






\maketitle

\begin{abstract}
The rise of reinforcement learning (RL) in critical real-world applications demands a fundamental rethinking of privacy in AI systems. 
Traditional privacy frameworks, designed to protect isolated data points, fall short for sequential decision-making systems where sensitive information emerges from temporal patterns, behavioral strategies, and collaborative dynamics.
Modern RL paradigms, such as federated RL (FedRL) and RL with human feedback (RLHF) in large language models (LLMs), 
exacerbate these challenges by introducing complex, interactive, and context-dependent learning environments that traditional methods do not address.
In this position paper, we argue for a new privacy paradigm built on four core principles: multi-scale protection, behavioral pattern protection, collaborative privacy preservation, and context-aware adaptation. 
These principles expose inherent tensions between privacy, utility, and interpretability that must be navigated as RL systems become more pervasive in high-stakes domains like healthcare, autonomous vehicles, and decision support systems powered by LLMs. 
To tackle these challenges, we call for the development of new theoretical frameworks, practical mechanisms, and rigorous evaluation methodologies that collectively enable effective privacy protection in sequential decision-making systems.

\end{abstract}


\section{Introduction}\label{sec:introduction}
The rise of reinforcement learning (RL) in critical real-world applications \cite{Mnih2015,Sutton2018,Dulac-Arnold2021,corecco2024suber,lu2023action} has exposed a fundamental tension in AI privacy: How do we protect sensitive information in systems that learn and make decisions over time? Traditional privacy frameworks, built for protecting individual data points in static datasets \cite{Dwork2006,Abadi2016,Papernot2016}, are increasingly inadequate for modern RL systems where sensitive information exists not just in individual moments but in temporal patterns, behavioral strategies, and collaborative dynamics \cite{gomrokchi2023membership}. These privacy challenges arise directly from RL's fundamental characteristic of learning through sequential interaction.

\begin{figure}[t]
    \centering
    \includegraphics[width=\linewidth]{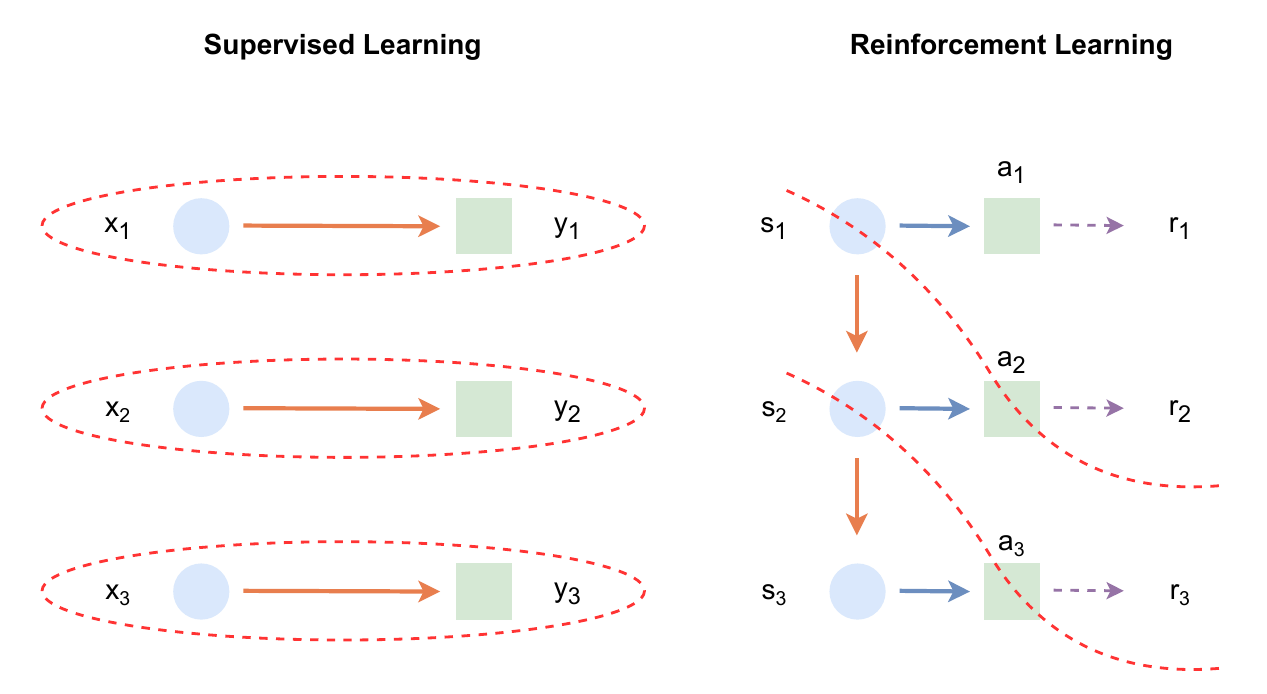}
\caption{
\textbf{The essential contrast in privacy requirements between supervised learning (left) and reinforcement learning (right)}. In supervised learning, data points $(x,y)$ are isolated and can be protected individually through local privacy mechanisms (red dashed ellipses). In reinforcement learning, a sequence of states $s$ connected by actions $a$ and rewards $r$ creates temporal dependencies, highlighted by the red dashed curve that spans multiple decision points. While local privacy protection can be adapted in reinforcement learning, the sequential nature of state transitions and action-reward pairs creates dependencies that make point-wise privacy protection insufficient.}
    \label{fig:sequential_privacy}
\end{figure}

Unlike traditional machine learning paradigms, reinforcement learning operates through continuous interaction between an agent and its environment \cite{Sutton2018}. The RL agent observes the current state of the environment, takes actions based on these observations, and receives feedback in the form of rewards. This sequential learning process fundamentally differs from supervised learning, where most existing privacy frameworks were developed \cite{Dwork2006, Abadi2016}. As illustrated in Fig.~\ref{fig:sequential_privacy}, supervised learning data points are typically treated as independent samples, allowing privacy mechanisms to protect each point individually. However, RL violates this independence assumption, giving rise to three fundamental privacy concerns: First, \emph{temporal patterns} emerge from sequential relationships between states, actions, and rewards, creating dependencies that span entire trajectories and potentially revealing sensitive information about the underlying process \cite{Ma2019}. Second, \emph{behavioral strategies} develop as the agent learns optimal policies, encoding complete decision-making patterns that go beyond the simple input-output mappings of supervised learning and may reveal proprietary algorithms or institutional expertise \cite{Cundy2024}. Third, \emph{collaborative dynamics} arise from the continuous adaptation between the agent and its environment, creating ongoing relationships that have no parallel in traditional supervised learning and potentially exposing sensitive information through patterns of response and adjustment \cite{Hitaj2017}.  

While these privacy challenges are fundamental to basic RL systems, the emergence of advanced paradigms has further amplified these concerns, particularly in relation to collaborative dynamics. Federated reinforcement learning (FedRL), where multiple agents share learning experiences while keeping data locally \cite{Fan2021,fan2023fedhql,Woo2023,Jordan2024,yue2024momentum,Di2024FedRL,Woo2024,jiang2025fedhpd}, introduces the challenge of protecting not only individual agent data but also emergent collective behavioral patterns. Similarly, large language models (LLMs), such as ChatGPT~\cite{ChatGPT2023} and DeepSeek~\cite{Deepseek2024}, refined through reinforcement learning with human feedback (RLHF) \cite{Christiano2017,Ziegler2019,Stiennon2020} extend these collaborative privacy challenges to human-AI interaction, creating additional vulnerabilities around protecting annotator characteristics and cultural information encoded in feedback patterns~\cite{corecco2024llm,fan2024fedrlhf}.
Recent analysis has uncovered numerous instances of personal data in publicly available RLHF datasets that had evaded removal \cite{von2024cannot}, highlighting how even carefully curated training data can expose private user information.
These advanced paradigms demonstrate how the temporal and behavioral aspects of privacy intertwine with collaborative dynamics, pushing privacy challenges beyond individual agent privacy to encompass group-level patterns and societal concerns.


Recent privacy regulations like GDPR \cite{GDPR} and HIPAA \cite{HIPAA} establish strict requirements for protecting such sensitive information, but their frameworks—designed primarily for static data protection—struggle to address these dynamic aspects of RL systems. These regulations presume a clear distinction between protected and non-protected data, a distinction that blurs in RL systems where sensitive information often emerges from patterns of interaction rather than residing in individual data points. This fundamental mismatch between regulatory frameworks and the nature of RL systems creates significant challenges for deployment in regulated domains.

To address these challenges, this position paper:
\begin{enumerate}
    \item Articulates why traditional privacy frameworks fundamentally fail for sequential decision-making systems
    \item Proposes four core principles for a new privacy paradigm: multi-scale protection, behavioral pattern protection, collaborative privacy preservation, and context-aware adaptation
    \item Identifies critical open problems and research directions for realizing effective privacy in sequential settings
\end{enumerate}
The rest of this paper is organized as follows: Section~\ref{sec:evolution_of_privacy} reviews the evolution of privacy approaches in sequential settings. Section~\ref{sec:why_traditional_privacy_fail} explains why traditional frameworks fail. 
Section~\ref{sec:core_principles} proposes four core principles that leads to our \emph{Sequential Privacy} framework for rethinking privacy in RL setting. 
Section~\ref{sec:sequential_privacy_in_practice} examines implications through real-world applications. 
Section~\ref{sec:research_directions} outlines research directions, and 
we conclude in Section~\ref{sec:conclusion} with a call for community action toward developing privacy frameworks that can meet the needs of modern RL systems.

\section{Evolution of Privacy Approaches in Sequential Settings}\label{sec:evolution_of_privacy}

Before examining why traditional approaches fail, we trace the historical development of privacy mechanisms and their attempts to address sequential decision-making contexts. This evolution reveals a progression of increasingly sophisticated approaches, each trying to overcome the limitations of its predecessors while inadvertently highlighting deeper challenges.

\subsection{Technical Foundations}
The field of privacy-preserving machine learning began with differential privacy (DP), introduced by Dwork et al. \cite{Dwork2006}. This framework provides a mathematical foundation for quantifying information leakage: a randomized mechanism $\mathcal{M}$ satisfies $(\epsilon,\delta)$-differential privacy if changes to individual data points have only limited impact on the output distribution. Formally:

For any two \emph{adjacent} datasets $D$ and $D'$ (differing in at most one data record), and for all measurable subsets $S$ of possible outputs, $\mathcal{M}$ satisfies $(\epsilon,\delta)$-DP if:
\[
\Pr[\mathcal{M}(D) \in S] \;\le\; e^{\epsilon}\,\Pr[\mathcal{M}(D') \in S]\;+\;\delta.
\]
Here, $\epsilon$ (the ``privacy budget'') controls the multiplicative gap in probabilities, while $\delta$ bounds the probability of a larger deviation. Smaller $\epsilon$ and $\delta$ imply stronger privacy guarantees.

Two properties made this framework particularly attractive: 
\begin{itemize}
    \item \emph{Composition}: Privacy guarantees combine predictably over multiple analyses of the same dataset.
    \item \emph{Post-processing}: Privacy guarantees persist under any data-independent transformation of the mechanism’s output.
\end{itemize}
These properties proved highly effective for static data but would later reveal fundamental limitations in sequential settings.

\subsection{Early Adaptations to Sequential Data}
The first attempts to apply privacy to sequential settings emerged in the early 2010s, as researchers began working with temporal data like reinforcement learning trajectories:
\[
\tau = (s_0, a_0, r_0, s_1, a_1, r_1,\ldots,s_T)
\]
Early approaches split into two camps: those applying noise independently at each timestep, and those treating entire trajectories as atomic units. Both approaches revealed concerning trade-offs between privacy guarantees and system utility. Building on the original DP framework, later work extended the ideas to streaming and continual observation settings \cite{Dwork2010,Dwork2010b} and further explored private sequential learning in online contexts \cite{Tsitsiklis2018} and with bandit feedback \cite{Agarwal2017}. These efforts highlighted a fundamental tension: the very temporal correlations that make sequential learning effective also create new privacy vulnerabilities.

\subsection{The Cryptographic Era}
As limitations of noise-based approaches became apparent, the field of privacy-preserving  learning shifted toward cryptographic solutions. Researchers explored secure multi-party computation and homomorphic encryption, aiming to enable secure computation without data sharing~\cite{Sakuma2008}. 
At the same time, privacy-preserving deep learning emerged, with early frameworks showing how collaborative deep models could be trained without directly sharing sensitive data \cite{Shokri2015,Gilad2016}. While these approaches provided strong cryptographic guarantees, they faced significant scalability challenges and, more fundamentally, couldn't address the broader issue of behavioral pattern privacy that emerges in sequential settings~\cite{Hitaj2017}.

\subsection{Information-Theoretic Approaches}
The late 2010s saw researchers turn to information theory, introducing mutual information constraints to limit information leakage through learned policies \cite{Cuff2016,Fredrikson2015}. This marked an important shift in thinking—from protecting individual data points to considering the information content of behavioral patterns. While these approaches provided new theoretical insights, they highlighted the difficulty of balancing privacy with the need to preserve useful temporal patterns.

\subsection{Modern Developments}
Recent years have seen two parallel developments that further complicate the privacy landscape. On one front, advances in deep learning under differential privacy have been refined and deployed at scale \cite{Abadi2016,Apple2017}. These works leverage advanced privacy accounting (e.g., the moment accountant) to tightly track cumulative privacy loss during iterative training, ensuring high utility despite strict privacy constraints. On another front, theoretical insights such as privacy amplification by iteration have demonstrated that the effective privacy loss in iterative algorithms can be significantly reduced \cite{Feldman2018}. Moreover, modern private learning approaches continue to grapple with the challenges of gradient inversion and recovering sensitive information from model updates \cite{Zhu2019,Fredrikson2015}.

\section{Why Traditional Privacy Approaches Fail}\label{sec:why_traditional_privacy_fail}

The historical evolution of privacy approaches reveals not just technical limitations but fundamental incompatibilities with sequential decision-making. Here we analyze why these approaches fail, showing that the challenges arise from core properties of sequential learning rather than implementation limitations.

\subsection{The Sequential Nature of RL}
Consider the structure of RL data: a continuous stream of interactions between an agent and its environment, generating trajectories of the form
\[
\tau = (s_0, a_0, r_0, s_1, a_1, r_1,\ldots,s_T)
\]
where each state-action pair influences the entire future sequence. As shown in Fig.~\ref{fig:sequential_privacy}, unlike supervised learning data, which is often (implicitly) assumed to be independent and identically distributed, RL trajectories exhibit strong temporal dependencies~\cite{balle2016differentially}. 

Because decisions at each timestep impact future states and rewards, RL reveals sensitive information not simply at isolated moments but in multi-step patterns and dynamic behaviors. As a result, privacy threats can emerge in unforeseen ways. For example, small changes in an action sequence can cascade and expose strategic aspects of a policy, or aggregated trajectories across agents can leak private information about a collective population. These challenges do not arise merely from implementation details; rather, they stem from the fundamental \emph{sequential} nature of RL.

\begin{table*}[t]
\caption{Summary of Core Privacy Challenges in Sequential Decision-Making}
\label{tab:privacy_challenges}
\centering
\begin{tabular}{p{2.5cm}p{4cm}p{4cm}p{6cm}}
\hline
\textbf{Privacy Challenge} & \textbf{Root Cause} & \textbf{Prior Work Attempts} & \textbf{Limitations} \\
\hline
Temporal & Multi-step correlations across trajectory segments & \cite{Dwork2010}, \cite{Dwork2010b}, \cite{Zhang2022} & Standard composition theorems insufficient; privacy loss accumulates faster than expected \\
\hline
Behavioral & Learned policy encodes sensitive information in behavioral patterns & \cite{Cundy2024}, \cite{gomrokchi2023membership} & Per-sample protection fails to bound policy-level leakage; dynamic sensitivity undermines guarantees \\
\hline
Collaborative & Non-local information flow across agents and human feedback & \cite{Hitaj2017}, \cite{Zhu2019}, \cite{fan2024fedrlhf} & Local protection mechanisms ignore global patterns; aggregated updates expose institutional strategies \\
\hline
Context-Dependent & Domain-specific constraints and varying regulatory requirements & \cite{SAE2018}, \cite{GDPR}, \cite{HIPAA}, \cite{Karnouskos2017} & One-size-fits-all frameworks inadequate for diverse deployment contexts \\
\hline
\end{tabular}
\end{table*}

\subsection{Exploration-Exploitation Privacy Vulnerabilities}
The exploration-exploitation dilemma—central to RL but absent in supervised learning—introduces additional privacy concerns~\cite{zhao2024privacy}. During exploration, an agent must try different actions to discover optimal strategies, creating behavioral patterns that can leak sensitive information about:

\begin{itemize}
    \item \emph{Knowledge boundaries}: Exploration patterns reveal what an agent doesn't know, potentially exposing gaps in training data
    \item \emph{Learning dynamics}: The transition from exploration to exploitation creates temporal signatures that reflect training procedures
    \item \emph{Uncertainty profiles}: Methods using uncertainty-based exploration (e.g., UCB algorithms) directly expose confidence estimates derived from private data
\end{itemize}

As a consequence, traditional privacy mechanisms face a new challenge: adding sufficient noise to mask exploration patterns can severely impair learning efficiency, while preserving learning efficiency may reveal sensitive information through exploration behavior. This tension exacerbates the already complex privacy-utility tradeoff in RL.

\subsection{Analysis of Core Limitations}\label{sec:analysis_challenges_limitations}
Building on the above observations, we identify four core limitations that make traditional privacy approaches fundamentally inadequate for RL systems. 

\subsubsection{Temporal Privacy Challenge}
\label{sec:temporal-challenge}
The \emph{Temporal Privacy Challenge} arises from the fact that each timestep in an RL trajectory is tightly coupled with past and future timesteps. As shown by Mironov~\cite{Mironov2017} and Zhang et al.~\cite{Zhang2022}, privacy loss can grow faster than standard composition theorems would predict, because small inferences at one timestep accumulate into bigger insights about the entire trajectory. In other words, even if each individual $(s_t, a_t)$ pair is “protected” in isolation, long-range temporal correlations can still reveal private information. This challenge implies that privacy cannot be guaranteed by simply masking individual points; we must consider correlations across multiple temporal scales.

\subsubsection{Behavioral Privacy Challenge}
\label{sec:behavioral-challenge}
The \emph{Behavioral Privacy Challenge} stems from \emph{dynamic sensitivity} and the possibility that a learned policy encodes sensitive information in its behavioral patterns \cite{pan2019you,gomrokchi2023membership}. A shift in a single action can propagate through future states and unravel details about the underlying policy (or the environment that shaped that policy). Cundy et al.~\cite{Cundy2024} show how observing an agent’s behavior distribution can leak crucial details about training data or proprietary algorithms. Traditional privacy frameworks that focus on per-sample data protection cannot bound this “policy-level” leakage. In domains where the strategy itself is sensitive (e.g., proprietary approaches in healthcare or competitive settings), such behavioral leakage is a direct threat.

\subsubsection{Collaborative Privacy Challenge}
\label{sec:collab-challenge}
The \emph{Collaborative Privacy Challenge} emerges from multi-agent systems, federated reinforcement learning (FedRL), and human-in-the-loop methods (RLHF). These settings involve continuous sharing of updates, observations, or feedback among different parties (agents, servers, humans). Even if each local dataset or feedback instance is protected, the \emph{aggregated patterns} of interaction and adaptation can reveal sensitive information about individuals or institutions~\cite{Hitaj2017, Zhu2019}. This non-local information flow is especially difficult to secure, because high-level behaviors or group updates can expose private attributes that traditional pointwise protection ignores.

\subsubsection{Context-Dependent Privacy Challenge}
\label{sec:context-challenge}
Finally, the \emph{Context-Dependent Privacy Challenge} arises from the fact that privacy sensitivities and regulatory requirements vary drastically across different domains, user populations, and deployment contexts. A healthcare RL system must comply with HIPAA or GDPR, while an autonomous vehicle system may face different proprietary or safety-driven constraints \cite{SAE2018}. Similarly, an LLM refined via RLHF may need to guard the cultural or demographic information of human annotators in ways that differ from other RL use cases. Privacy is not “one size fits all,” and the severity of temporal, behavioral, and collaborative leakage depends intimately on the context in which the RL system is deployed. This challenge calls for \emph{adaptive} privacy frameworks that respond to domain- or population-specific requirements.

\subsection{Failed Extension Attempts}
Attempts to extend traditional privacy frameworks to RL reveal fundamental limitations. Adding noise at individual timesteps often destroys sequential structure, while protecting entire trajectories sacrifices utility. Cryptographic approaches~\cite{Sakuma2008} provide strong guarantees but face prohibitive computational overhead and still struggle with higher-level behavioral patterns.

Sophisticated techniques in federated or collaborative settings cannot fully prevent leakage of global patterns, as demonstrated by gradient inversion attacks~\cite{Zhu2019} and adaptive interactions revealing participant attributes~\cite{Hitaj2017}. Recent work~\cite{Rajabi2022,Vietri2020} shows progress under restricted settings but still fails to address the interplay of temporal, behavioral, collaborative, and contextual factors.

\subsection{Implications for Privacy Design}
\label{sec:implications}

Taken together, these observations emphasize that privacy in RL cannot be retrofitted with simple modifications of traditional frameworks. The four challenges require us to:

\begin{enumerate}
    \item \emph{Address multi-step correlations} (temporal challenge) $\rightarrow$ \textbf{Multi-scale Privacy Protection} (Section~\ref{sec:principles_multi_scale_privacy})
    \item \emph{Protect the policy itself} (behavioral challenge) $\rightarrow$ \textbf{Behavioral Pattern Protection} (Section~\ref{sec:principles_behavioral_pattern_privacy})
    \item \emph{Preserve privacy across interacting agents and humans} (collaborative challenge) $\rightarrow$ \textbf{Collaborative Privacy Preservation} (Section~\ref{sec:principles_collaborative_privacy})
    \item \emph{Adapt protections to domain-specific constraints} (context-dependent challenge) $\rightarrow$ \textbf{Context-Aware Adaptation} (Section~\ref{sec:principles_context_aware_privacy})
\end{enumerate}

Any new approach must \emph{jointly} tackle these dimensions while striking a careful balance between privacy, utility, interpretability, and feasibility. These requirements motivate the principles we propose next.

\section{Core Principles for Sequential Privacy}
\label{sec:core_principles}

We now introduce four core principles that address the challenges identified in Section~\ref{sec:why_traditional_privacy_fail}.


\subsection{Multi-scale Privacy Protection}
\label{sec:principles_multi_scale_privacy}

To counteract the \emph{Temporal Privacy Challenge}, privacy must hold across multiple temporal scales---not just at the granularity of individual actions or states. We extend traditional definitions of differential privacy to account explicitly for trajectory segments of varying lengths~\cite{Mironov2017,Zhang2022}:

\begin{definition}[Multi-scale Privacy]\label{def:multi_scale_privacy}
A mechanism $\mathcal{M}$ provides $(k,\epsilon,\delta)$-multi-scale privacy if for all scales $1 \le j \le k$ and all trajectory segments $\tau_{t:t+j}$,
\[
\Pr[\mathcal{M}(\tau_{t:t+j})\in S]\;\le\; e^{\epsilon_j}\,\Pr[\mathcal{M}(\tau'_{t:t+j})\in S]\;+\;\delta_j,
\]
where $\tau_{t:t+j}, \tau'_{t:t+j}$ are adjacent trajectory segments of length $j$, and $\epsilon_j$ (respectively $\delta_j$) may increase with segment length.
\end{definition}

Such multi-scale protection ensures that RL trajectories do not leak information cumulatively over time, thereby limiting the adversary’s ability to reconstruct sensitive patterns from sequential data.
It generalizes the usual composition theorems in differential privacy \cite{Dwork2006} by allowing an $\epsilon_j$ budget for each trajectory segment length $j$. 
In practice, one might set $\epsilon_j$ to grow sub-linearly in $j$ to reflect partial reuse of noise across overlapping segments, or adopt advanced composition results that limit how quickly the overall privacy budget depletes over time. A rigorous analysis requires bounding correlations between overlapping segments $\tau_{t: t+j}$, which is an open theoretical question. For instance, one could assume a Markov property and then derive $\epsilon_j$ by combining concentration inequalities with standard DP composition results--however, the exact rate of growth in $\epsilon_j$ would depend on the mixing time of the underlying Markov chain. Investigating these parameter choices remains an important research direction.

\subsection{Behavioral Pattern Protection}
\label{sec:principles_behavioral_pattern_privacy}

Addressing the \emph{Behavioral Privacy Challenge} requires protecting the \emph{policy}---i.e., the mapping from states to actions---rather than just the individual samples. This protection must cover both \emph{exploitation patterns} (revealing what was learned) and \emph{exploration patterns} (revealing uncertainty and learning dynamics). We thus focus on bounding divergences between entire trajectory distributions induced by different policies~\cite{Cundy2024}:

\begin{definition}[Behavioral Pattern Privacy]\label{def:behavioral_pattern_privacy}
A policy learning mechanism $\mathcal{M}$ satisfies $(\alpha,\beta)$-behavioral privacy if for any policies $\pi_1, \pi_2$ learned from adjacent training sets,
\[
D_{\alpha}\!\big(\mathbb{P}_{\tau\sim\pi_1}\,\Vert\,\mathbb{P}_{\tau\sim\pi_2}\big)\;\le\;\beta,
\]
where $D_{\alpha}$ is the R\'enyi divergence~\cite{Mironov2017} of order $\alpha$ and $\mathbb{P}_{\tau\sim\pi}$ is the distribution of trajectories under policy $\pi$.
\end{definition}

This definition encompasses both exploitation and exploration behaviors, as $\pi$ contains both components. However, exploration patterns present unique privacy challenges: they directly reveal uncertainty estimates which are typically derived from training data distributions. For example, in algorithms using upper confidence bounds (UCB) or Thompson sampling, the exploration strategy directly exposes confidence intervals calculated from private data.

By bounding how much a single agent's (or institution's) policy distribution—including both exploitation and exploration components—can shift under small changes in the underlying data, we reduce the risk that adversaries infer proprietary strategies, specialized treatment protocols, data sparsity patterns, or other policy-level knowledge.

\subsection{Collaborative  Privacy Preservation}
\label{sec:principles_collaborative_privacy}

The \emph{Collaborative Privacy Challenge} is especially evident in federated or multi-agent RL, and in RLHF where human feedback is continuously integrated. We can frame this via information-theoretic constraints on collaborative systems~\cite{Cuff2016}:

\begin{definition}[Collaborative Privacy]\label{def:collaborative_privacy}
A mechanism $\mathcal{M}$ provides $(\gamma,\eta)$-collaborative privacy if for all interaction histories $H_t$ and new interactions $i_t$:
\[
I(\mathcal{M}(H_t \cup \{i_t\}); \text{sensitive}_t | H_t) \leq \gamma
\]
with probability at least $1-\eta$, where $I(\cdot;\cdot|\cdot)$ denotes conditional mutual information and $\text{sensitive}_t$ represents any sensitive attributes at time $t$ including:
\begin{itemize}
    \item Demographic information of participants
    \item Group-level behavioral patterns
    \item Institutional strategies or protocols
    \item Collective learning dynamics
\end{itemize}

\end{definition}

This definition limits how much \emph{additional} information is revealed about sensitive attributes (e.g., user demographics, group-level strategies) from each incremental interaction, even when prior interactions are already known.

\subsection{Context-Aware Adaptation}
\label{sec:principles_context_aware_privacy}

Finally, the \emph{Context-Dependent Privacy Challenge} demands that privacy guarantees adapt to different domains, user populations, and regulatory environments. We capture this adaptivity via:

\begin{definition}[Context-Aware Privacy]\label{def:context_aware_privacy}
A privacy mechanism $\mathcal{M}$ is $(\Theta,\lambda)$-context-aware if for all contexts $c \in \mathcal{C}$ and privacy requirements $\theta_c \in \Theta$:
\[
\Pr\big[\mathcal{M}(\tau,c) \text{ satisfies } \theta_c\big] \;\geq\; 1-\lambda,
\]
where $\theta_c$ specifies context-specific privacy parameters.
\end{definition}

In high-stakes environments (e.g., clinical healthcare), $\theta_c$ may demand stricter bounds and narrower noise budgets, while less-sensitive tasks can tolerate relaxed protections. The mechanism adjusts its privacy parameters or noise injection strategies according to these evolving contextual requirements.
It formalizes the idea that privacy mechanisms must adapt to different contexts while maintaining guaranteed levels of protection \cite{Papernot2016}.

For instance, in a hospital environment, context-aware privacy might mean enforcing tighter privacy budgets $(\epsilon)$ for particularly sensitive patient attributes, in compliance with HIPAA or GDPR, while still allowing less-protected telemetry data to facilitate real-time decision-making. 
By contrast, in autonomous vehicles, the context might revolve around location data and proprietary driving logs: the system could relax certain bounds for purely operational metrics (e.g., mechanical sensors), but apply stricter protections for route data or user identities. These domain-specific variations underscore why static, one-size-fits-all privacy mechanisms often fail in practice: each context demands unique trade-offs between privacy, regulatory compliance, and system performance.

\subsection{The Sequential Privacy Framework}\label{sec:principles_unified_framework}

These four principles form our
\emph{Sequential Privacy Framework} for reinforcement learning in sequential decision-making. A straightforward way to ensure that all aspects of privacy are satisfied is to enforce each principle independently and then take a worst-case (intersection) view of the guarantees. Formally, the overall privacy guarantee can be expressed as:
\[
\mathcal{P}(\tau) = \min_{i\in\{1,2,3,4\}} \mathcal{P}_i(\tau),
\]
where $\mathcal{P}_i(\tau)$ represents the privacy guarantee derived from each of the four principles (multi-scale, behavioral, collaborative, and context-aware). This worst-case perspective provides a conservative baseline, ensuring that an adversary cannot exploit any single dimension of leakage.

In practice, implementing these principles may involve combining multiple mechanisms:
\[
\mathcal{M}(\tau) = h\!\big(
  \mathcal{M}_{\text{multi}}(\tau),\;
  \mathcal{M}_{\text{behav}}(\tau),\;
  \mathcal{M}_{\text{collab}}(\tau),\;
  \mathcal{M}_{\text{context}}(\tau)
\big),
\]
where each $\mathcal{M}_{\text{multi}},\mathcal{M}_{\text{behav}},\mathcal{M}_{\text{collab}},\mathcal{M}_{\text{context}}$ enforces one of the four privacy principles, and $h$ composes their outputs or noise parameters. For example, $\mathcal{M}_{\text{multi}}$ might add calibrated noise to gradient updates at multiple timescales, while $\mathcal{M}_{\text{behav}}$ further imposes policy-level divergence bounds to ensure an entire learned policy does not reveal sensitive information. The function $h$ could be a higher-level controller that orchestrates noise or post-processing across these components, balancing their respective privacy-utility trade-offs.

We note, however, that existing composition theorems for differential privacy are largely tailored to static or i.i.d.\ settings~\cite{Dwork2006}, and directly applying them in sequential RL may be over-restrictive or suboptimal. Investigating novel composition rules that account for temporal dependence, policy-level constraints, and collaborative feedback remains a key open research problem. Future work could explore alternative ways of combining these principles to yield a single privacy budget, or develop {contextual composition} strategies that selectively apply stricter bounds in high-risk scenarios.

\subsection{Theoretical Bounds on Privacy-Utility Trade-offs}
\label{sec:theoretical_bounds}

Our Sequential Privacy framework implies that any mechanism satisfying
\((\alpha,\beta)\)\nobreak\emph{–behavioral privacy} must incur a quantifiable performance cost.
Below we state a concrete bound under standard finite‑MDP assumptions.

\begin{lemma}[Privacy–Utility Trade‑off (Sketch)]\label{thm:privacy_utility}
In any finite MDP with discount $\gamma$, a mechanism satisfying 
$(\alpha,\beta)$–behavioral privacy (Def.~4.2) must incur
\[
  \mathbb{E}\bigl[V^{\pi^*}(s_0) - V^{\pi_{\rm priv}}(s_0)\bigr]
  \;=\;\Omega\!\bigl((1-\gamma)/\beta\bigr).
\]
\end{lemma}

The above results can be obtained by applying the standard performance-difference lemma \cite{kakade2002approximately} to the R\'enyi bound. It mirrors the familiar inverse‑scaling
trade‑offs in differentially‐private supervised learning (e.g.\
\(\Omega(1/\epsilon)\) lower bounds in DP‑SGD \cite{bassily2014private}),
but here it applies at the level of entire \emph{policies} rather than per‑step
gradient updates.  It concretely demonstrates that \emph{any} mechanism
strongly limiting policy divergence must pay a nontrivial price in
expected return.




\section{Sequential Privacy in Practical Applications}
\label{sec:sequential_privacy_in_practice}

Our Sequential Privacy framework addresses critical challenges in high-stakes domains. Here we demonstrate how the principles can be implemented in three key areas.

\subsection{Healthcare: Privacy-Critical Treatment Optimization}
RL systems optimizing treatment strategies must protect temporal patterns in patient care that could reveal both individual conditions and institutional protocols. For chronic conditions like diabetes, blood glucose and insulin adjustment sequences encode sensitive information even when individual decisions are protected.
%
In such clinical settings, policy gradient methods can be adapted for multi-scale privacy:
\begin{align}
\theta_{t+1} = \theta_t + \eta \cdot \Big(\sum_{i} \text{clip}\left(\nabla_\theta \log \pi_\theta(a_i|s_i) A(s_i,a_i), C_{\text{med}}\right) \nonumber
\\ + \mathcal{N}(0, \sigma^2_{\text{med}} C^2_{\text{med}} \mathbf{I})\Big)\nonumber
\end{align}

Here, $C_{\text{med}}$ represents the HIPAA-compliant clipping threshold with $\sigma_{\text{med}}$ calibrated to satisfy differential privacy for sensitive medical attributes. Critical in healthcare is adaptive privacy budgeting, where diagnostic phases may receive stronger protection than maintenance phases while still preserving overall temporal pattern privacy.

\subsection{Autonomous Vehicles: Proprietary Strategy Protection}
Vehicle fleets generate massive behavioral data encoding navigation strategies and risk assessment algorithms. The behavioral pattern protection principle becomes crucial when companies share driving experiences to enhance safety while protecting proprietary algorithms~\cite{SAE2018,Karnouskos2017}.
%
Q-learning variants are particularly suited for autonomous vehicle settings, with behavioral pattern privacy implemented through:
\begin{align}
\pi(a|s) = \frac{\exp(Q(s,a)/\tau_{\text{auto}})}{\sum_{a'} \exp(Q(s,a')/\tau_{\text{auto}})} \nonumber
\end{align}

The temperature parameter $\tau_{\text{auto}}$ can be dynamically adjusted based on driving context—higher in routine navigation (preserving proprietary algorithms) and lower in safety-critical scenarios where precise behavior is essential. This implementation balances the need to share knowledge of hazardous scenarios while preserving competitive algorithmic advantages.

\subsection{LLMs: Human Feedback Privacy}
RLHF systems must protect not only model behavior but also the characteristics of human feedback providers~\cite{Carlini2021,fan2024fedrlhf}. Even with anonymized instances, preference patterns could reveal annotator demographics through temporal correlations.
%
For collaborative privacy in RLHF, we can implement a stratified protection approach:
\begin{align}
r_{\text{private}} = r_{\text{original}} + \text{Lap}(\Delta f / \epsilon_{\text{demo}}) \nonumber
\end{align}

where demographic-correlated feedback receives stronger protection ($\epsilon_{\text{demo}}$) than content-specific feedback. Critically, preference order preservation constraints must be maintained while obscuring demographic patterns. For federated RLHF settings, secure aggregation with temporal sensitivity weighting can further protect annotator characteristics while preserving useful preference signals.

\section{The Path Forward: A Research Agenda}
\label{sec:research_directions}

The preceding sections highlight critical questions for achieving robust
privacy in sequential decision-making. Here, we sketch four complementary
directions
that together form a research agenda for
\emph{sequential privacy} in RL.

\subsection{Theoretical Foundations}
While classical differential privacy provides a strong
baseline in static settings, \emph{sequential} RL poses unique challenges
due to overlapping trajectories, temporal dependencies, and adaptive
interaction. Researchers must formalize privacy notions specifically for
these correlated settings, extending composition theorems to account for
overlapping segments or multi-scale observations. For instance, bounding
privacy leakage at partial trajectory segments and analyzing privacy
amplification under Markovian assumptions remain open problems. Establishing
\emph{impossibility} results—where no mechanism can simultaneously achieve
strong privacy and high utility for certain classes of RL tasks—would also
offer valuable theoretical guidance. Lastly, rigorous empirical metrics
are needed to quantify privacy-utility trade-offs \emph{across} different
time horizons.

\subsection{Mechanism Design for Temporal Privacy}
Existing privacy approaches in RL typically protect either individual
timesteps or entire trajectories, risking either excessive noise or
unmitigated leakage. Future work must blend \emph{adaptive noise injection}
and \emph{multi-scale perturbations} so that data at highly sensitive
timesteps is masked more heavily, while allowing enough signal to learn
effective policies. Policy-level regularization methods—such as constraining
the divergence between learned policies and a reference policy—could further
limit the risk of revealing private information through policy behaviors.
Additionally, designing lightweight, domain-aware privacy layers for
continuous or partially observed environments would expand the applicability
of privacy-preserving RL beyond discrete, small-scale benchmarks.

\subsection{Collaborative Privacy Preservation}
Federated RL, multi-agent RL, and RLHF settings introduce continuous
coordination and real-time feedback among agents or human annotators. Classic individual-level privacy guarantees (e.g.,
per-user DP) often fail to capture \emph{group-level} or cross-party
inferences. New metrics are thus needed to quantify information leakage
in group updates or shared gradients, and mechanisms must ensure that
aggregated model parameters do not inadvertently reveal \emph{collective}
sensitive patterns. In RLHF scenarios, privacy solutions must conceal
annotator identities and attributes, even as the model iteratively
incorporates feedback to refine its policies. Developing robust defenses
against membership inference, gradient inversion, and other adaptive
attacks in collaborative RL is paramount for real-world trust.

\subsection{Implementation and Deployment}
Bridging theory and practice requires tools for measuring privacy leakage
and ensuring scalable algorithmic performance in complex RL tasks. This
includes (1) developing standardized benchmarks and simulations that
stress-test privacy mechanisms under diverse temporal structures, (2)
creating open-source software libraries that integrate privacy-by-design
principles into typical RL pipelines (e.g., policy gradient or Q-learning
frameworks), and (3) defining domain-specific best practices to satisfy
regulatory or ethical constraints in sensitive environments such as
healthcare or autonomous driving. Ultimately, practical deployment
necessitates reconciling privacy with real-world demands for minimal
latency, interpretability, and fault-tolerance, underscoring the need
for multi-disciplinary collaboration among ML researchers, domain experts,
and policymakers.

\section{Conclusion}\label{sec:conclusion}

Reinforcement learning has rapidly evolved from a research frontier
to a technology shaping critical real-world applications in healthcare,
transportation, and AI services like language models. 
Yet existing privacy frameworks,
designed primarily for static, pointwise data protection, leave these
sequential systems vulnerable. 
As we have illustrated, privacy breaches
in RL can reveal not only isolated data points but entire temporal or
behavioral strategies, along with emergent insights about collaborating
agents and their contexts.

To address these challenges, we introduce the \emph{Sequential Privacy} framework
built on four fundamental principles: multi-scale protection, behavioral pattern
protection, collaborative preservation, and context-aware adaptation.
Delivering on this vision demands new theory for
temporal and group‑level privacy, domain‑aware mechanisms, and standardized
evaluations that balance privacy, utility, and interpretability.

The time is ripe to confront the inseparable link between sequential
decision-making and emergent privacy risks. By building on the \emph{Sequential Privacy} 
principles and open research questions we have posed, the broader AI community can
foster a more secure and privacy-preserving foundation for the next
generation of reinforcement learning systems.

\section*{Acknowledgment}

This research is supported by National Research Foundation, Singapore and Infocomm Media Development Authority under its Trust Tech Funding Initiative, the Centre for Frontier Artificial Intelligence Research, Institute of High Performance Computing, A*STAR, and the College of Computing and Data Science at Nanyang Technological University. Any opinions, findings, conclusions, or recommendations expressed in this material are those of the author and do not reflect the views of National Research Foundation, Singapore, and Infocomm Media Development Authority.

\newpage
\bibliographystyle{IEEEtran}
\bibliography{references}

\end{document}